\documentclass[conference]{IEEEtran}
\ifCLASSINFOpdf
  \usepackage[pdftex]{graphicx}
\else
\fi
\usepackage{amsmath}
\usepackage{subcaption}

\usepackage{booktabs}
\usepackage{multirow}
\usepackage[table,xcdraw]{xcolor}
\usepackage{listings}
\usepackage{colortbl}
\usepackage{makecell}
\usepackage{url}
\usepackage{enumitem}

\usepackage[utf8]{inputenc}
\usepackage{tcolorbox}
\usepackage{lipsum} 

\definecolor{codecomments}{rgb}{0,0.6,0}
\definecolor{codenumbers}{rgb}{0.5,0.5,0.5}
\definecolor{codestrings}{rgb}{.7, 0, .7}
\definecolor{codekeywords}{rgb}{0.8, 0.3, 0.4}

\lstdefinestyle{mystyle}{
    commentstyle=\color{codecomments},
    keywordstyle=\color{codekeywords},
    numberstyle=\tiny\color{codenumbers},
    stringstyle=\color{codestrings},
    basicstyle=\footnotesize\ttfamily,
    columns=flexible,
    breakatwhitespace=false,
    captionpos=b,
    keepspaces=true,
    showspaces=false,
    showstringspaces=false,
    showtabs=false,
    tabsize=2,
    frame=single
}

\usepackage{etoolbox}
\AtBeginEnvironment{quote}{\noindent\small\raggedright}
\renewenvironment{quote}
  {\list{}{\leftmargin=10pt \rightmargin=10pt}\item\relax\noindent\small\raggedright}
  {\endlist}

\begin{document}
%
\title{Explingo: Explaining AI Predictions using Large Language Models}



%
\author{\IEEEauthorblockN{Alexandra Zytek\IEEEauthorrefmark{1},
Sara Pido\IEEEauthorrefmark{1},
Sarah Alnegheimish\IEEEauthorrefmark{1}, 
Laure Berti-{É}quille\IEEEauthorrefmark{2} and
Kalyan Veeramachaneni\IEEEauthorrefmark{1}}
\IEEEauthorblockA{MIT\IEEEauthorrefmark{1}, Cambridge, MA\\
ESPACE-DEV IRD\IEEEauthorrefmark{2}, Montpellier, France\\
Email: zyteka@mit.edu}}


\maketitle

\begin{abstract}
Explanations of machine learning (ML) model predictions generated by Explainable AI (XAI) techniques such as SHAP are essential for people using ML outputs for decision-making. We explore the potential of Large Language Models (LLMs) to transform these explanations into human-readable, narrative formats that align with natural communication.
We address two key research questions: (1) Can LLMs reliably transform traditional explanations into high-quality narratives? and (2) How can we effectively evaluate the quality of narrative explanations? To answer these questions, we introduce \textsc{Explingo}, which consists of two LLM-based subsystems, a \textsc{Narrator} and \textsc{Grader}. The \textsc{Narrator} takes in ML explanations and transforms them into natural-language descriptions. The \textsc{Grader} scores these narratives on a set of metrics including accuracy, completeness, fluency, and conciseness. 

Our experiments demonstrate that LLMs can generate high-quality narratives that achieve high scores across all metrics, particularly when guided by a small number of human-labeled and bootstrapped examples. We also identified areas that remain challenging, in particular for effectively scoring narratives in complex domains. 
The findings from this work have been integrated into an open-source tool that makes narrative explanations available for further applications. 
\end{abstract}


%
\IEEEpeerreviewmaketitle

\newcommand{\note}[1]{\textcolor{red}{#1}}
\newcommand{\narrator}{narrator}
\newcommand{\Narrator}{\textsc{Narrator}}
\newcommand{\grader}{grader}
\newcommand{\Grader}{\textsc{Grader}}

\newcommand{\Explingo}{\textsc{Explingo}}

\newcommand{\goldstandard}{exemplar}
\newcommand{\GOLDSTANDARD}{EXEMPLAR}
\newcommand{\GoldStandard}{Exemplar}

\newcommand{\prompt}[1]{
    \begin{tcolorbox}[colback=gray!8, colframe=gray!80!black, width=\columnwidth, boxsep=1pt, left=-4pt, right=0pt, top=4pt, bottom=4pt]
    \noindent
    \begin{quote}
    #1
    \end{quote}
    \end{tcolorbox}
}

\section{Introduction}
\begin{figure*}
    \centering
    \includegraphics[width=\linewidth]{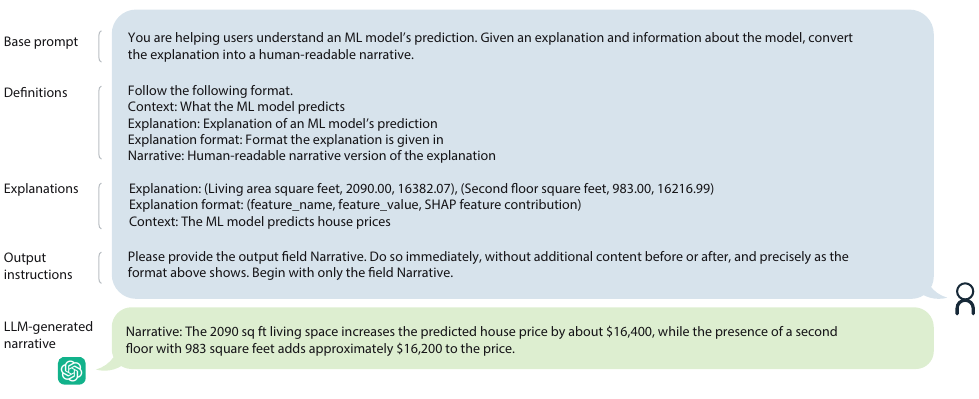}
    \caption{Sample \Narrator{} inputs and outputs. The items in blue make up the prompt passed to the \Narrator{} to transform ML explanations into narratives for a house-pricing example. The item in green is a narrative the \Narrator{} LLM may generate based on this prompt. Some components, like the output instructions, are provided by DSPy \cite{khattab2023dspycompilingdeclarativelanguage}.
    }
    \label{fig:narrator-prompt-simple}
\end{figure*}

Explanations of ML model predictions, such as those created by Explainable AI (XAI) algorithms like SHAP \cite{lundberg2017unified}, are important for making ML outputs usable for decision-making. It is difficult for users to make informed decisions using ML outputs without some explanation or justification of how ML model arrived at those. Past work \cite{lakkaraju_rethinking_2022, zytek_llms_2024} has suggested that some users may prefer explanations in a natural language format --- what we refer to in this paper as \textit{narratives}. The ``LLM-generated narrative'' in Fig. \ref{fig:narrator-prompt-simple} (green) shows an example of such a narrative. These narratives offer many benefits: they align with how humans naturally communicate, allow users to glean key takeaways quickly, and are a first step in enabling interactive conversations where users can better understand models by asking follow-up questions.

Large Language Models (LLMs) 
offer a promising way of generating such narratives.
Past work has suggested different ways to do so. Some work \cite{bhattacharjeeLLMguidedCausalExplainability2024, kroegerAreLargeLanguage2023} uses LLMs directly to explain ML predictions, without use of traditional XAI algorithms. While these approaches offer promising paths towards novel explanations, they also risk introducing inaccuracy or hallucination in explanations. An alternative approach, and the one we focus on in this paper, is to use LLMs to transform existing ML explanations generated using traditional, theoretically grounded XAI approaches into readable narratives. 

We seek to address two primary research questions:

\begin{enumerate}[left=0cm]
    \item Can LLMs effectively transform traditional ML explanations into high-quality narrative versions of explanations?
    \item How do we evaluate the quality of these narratives? 
\end{enumerate}

We address these questions with our two-part \Explingo{} system. To address the first question, we develop an LLM-based \Narrator{} that transforms SHAP explanations from diverse datasets into narratives. We then investigate what metrics may be of value to assess the quality of narratives, and auto-grade the generated narratives using our \Grader{} to address the second question. Automatically grading narratives is helpful not only for tuning and evaluating the system, but also for sure as a guardrail in deployment by hiding any narratives that are not of sufficient quality.

We note that users across different domains may have significantly different preferences for how these narratives should look. Some factors --- like the fact that narratives should accurately represent the original explanation and not be unnecessarily long --- are fairly universal. On the other hand, other factors are highly subjective. Some users may wish for a narrative that includes very precise and technical details, while others prefer more general information. 
In this paper, we aim to generate narratives that match the style of some small, user-provided \goldstandard{} dataset of narratives. Our objective is to create a system where, for any new application, a user can write a small number of sample narratives (around three to five) and get high-quality narrative versions of future ML predictions that mimic the linguistic style and structure of those samples.

Our key contributions from this work are:
\begin{itemize}
    \item We present our \Narrator{} system that transforms ML explanations into user-readable narratives.
    \item We develop an independent, automated \Grader{} system that assesses the quality of the generated narratives.
    \item We curated a sample set of nine \goldstandard{} datasets that contain narratives using four public datasets, that can be used for future narrative generation tuning and evaluation work. These datasets are public and can be accessed here: \url{https://github.com/sibyl-dev/Explingo}.
    \item We provide a comprehensive set of experiments to demonstrate the impact of different prompting and few-shot strategies on narrative generation, and evaluate potential sources of reduced narrative quality.
    \item Our system has been integrated in an open-source library here: \url{https://github.com/sibyl-dev/pyreal}.
\end{itemize}

\section{Background}
SHAP values \cite{lundberg2017unified} are a popular way to explain ML model predictions in terms of feature contributions. In these explanations, every feature is given a positive or negative value representing how much it increased or decreased the model's prediction on a specific input. Past work  has suggested these explanations are popular among users, but may be difficult for some user groups to parse \cite{zytek_sibyl_2021, suh_more_2024, bhatt_explainable_2020, jiang_two_2021, yang2023survey}. 

Past work has aimed to apply LLM's language-processing capabilities to XAI problems in different ways. Some work \cite{nguyen2023black, shenConvXAIDeliveringHeterogeneous2023, slack2023explaining} has focused on using LLMs at the input stage, taking in user questions and generating appropriate explanations to answer them. Other work \cite{bhattacharjeeLLMguidedCausalExplainability2024, kroegerAreLargeLanguage2023} has sought to use LLMs directly to generate explanations of model logic. 

In this paper, we instead build on work seeking to transform existing ML explanations into readable narratives. Burton et al. \cite{burton2023natural} used an LLM fine-tuned on a dataset of natural-language explanations to produce fluent and accurate narratives. Guo et al. \cite{guoAdaptingPromptFewshot2023} proposes a few-shot table-to-text generation approach that uses unlabeled domain-specific knowledge to convert tables into natural-language descriptions. We extend these works through an alternative scoring method based on a specific set of metrics, and use an approach based purely on a mix of manual and optimized prompting. 

Past work \cite{guo_evaluating_2023, bai_benchmarking_2023, chen_exploring_2023, ji_exploring_2023} has also looked into using LLMs to evaluate LLM outputs, in order to reduce effort and time required from a potentially small group of human users. Fu et al. \cite{fu_gptscore_2023} found that GPT-3 was able to effectively evaluate the quality of LLM-generated text based on a wide variety of metrics such as correctness, understandability, and diversity. Wang et al. \cite{wang_is_2023} similarly found that chatGPT is able to achieve state-of-the-art correlation with human graders. Liu et al. \cite{liu_g-eval_2023} found that advanced prompting techniques such as chain-of-thought prompting can further improve correlation between LLM and human grades. We applied these insights to optimize and evaluate our LLM-generated narratives using an LLM-based metric grader, eliminating the need for human graders and enabling evaluation across a wider range of domains without relying on domain experts.

\section{The \Explingo{} System: \Narrator{} and \Grader{}}
To transform XAI explanations into narratives, the \Explingo{} system includes two sub-systems:
1) \Narrator{}: a system to transform traditional explanations of ML model predictions into accurate, natural narratives.
2) \Grader{}: a metric grading system to assess the quality of narratives. 

\subsection{\Narrator{}: Transforming ML Explanations to Narratives}

Transforming traditional ML model explanations into readable narratives offers several potential benefits. First, these narratives allow key takeaways to be read quickly. Second, they are often the most natural format for parsing information for many users, who may prefer them to graphs or have difficulty understanding graphs that are not visualized effectively. Finally, offering narrative versions of explanations is the first step to allowing for full ``conversations'' with ML models, with users able to ask follow-up questions in natural-language and receive answers in kind.

The \Narrator{} is build on an LLM model, in our case GPT-4o, that is prompted to generate a narrative. The basic prompt is composed of four components: a base prompt, a definition of the input component structure, three specific explanation components (the ML explanation, the explanation format, and the context description), and output instructions. The prompt is then concluded with \textbf{Narrative:} invoking the LLM to start narrating. The first four blocks in Fig.~\ref{fig:narrator-prompt-simple} show an example prompt that follows this structure. As described in depth later, we additionally pass selected exemplar narratives that help the \Narrator{} customize its narratives for specific applications. 

The \textbf{ML explanation} --- for example, a SHAP feature contribution explanation, as we focus on in this paper --- is generated using some external library. We then parse this explanation into a consistent format, depending on the explanation type. For the SHAP explanation, we select the $N$ most contributing features by absolute value, and parse them in sorted order into the following format: 

\noindent\textit{(feature name 1, feature value 1, SHAP contribution 1), (feature name 2, feature value 2, SHAP contribution 2), \dots }

In addition to the explanation, the \narrator{} takes in a brief \textbf{explanation format} describing how to parse the given explanation. Passing this format description separately makes the system generalizable  to any kind of ML explanation that can be expressed in text. Moreover, it allows the explanation to be customized for additional information to be passed.

Finally, we write a brief description of the \textbf{context} of the explanation --- generally, what the ML model being explained is predicting. For SHAP explanations, this helps provide units to the contributions. 

The last box in Fig.~\ref{fig:narrator-prompt-simple} shows an example of the \Narrator{}'s output narrative.

Once the \Narrator{} generates narratives, we want to assess if these narratives are high-quality and expressing the right information. This is where the \Grader{} plays a role.

\begin{table*}[t]
\centering
\caption{Summary of \goldstandard{} datasets. In the Ames Housing Dataset \cite{de_cock_ames_2011}, each row represents one house, with 79 input features. The label to be predicted is the price the hold sold for. The Student Performance Dataset \cite{cortez_using_2008} includes 30 features, and each row represents a student. The label to be predicted is whether the child will pass or fail a class. The Mushroom Toxicity Dataset \cite{misc_mushroom_73} has 22 features about different specifies of mushrooms, and the predicted label is whether the mushroom is poisonous or not. Finally, the PDF Malware Dataset \cite{issakhani_pdf_2022} has 37 features about different PDFs, and the prediction label is whether the PDF contains malware. }
\label{tab:datasets}
\begin{tabular}{p{.07\linewidth}p{.14\linewidth}cp{.63\linewidth}}
\toprule 
ID & Dataset & \# of Entries & Sample Narrative \\
\midrule 
House 1    &  Ames Housing  & 35 &    The relatively smaller above ground living space in this house reduces its predicted price by about \$12,000. The lower-than-average material rating also reduced the house price by about \$10,000.         \\
House 2     &  Ames Housing & 22 &  The SHAP value indicates that the house price decreases because the above ground living area is 1256 sq ft and the material quality is rated 5.              \\
House 3     &   Ames Housing  & 22 &  This house is cheaper because it has less above ground living space (size=1256) and has lower material quality (rating=5).                \\
Mush 1     & Mushroom Toxicity & 30 &     This mushroom is more likely to be poisonous because its foul odor, silky stalk surface, and chocolate spore print color. Be careful!             \\
Mush 2     & Mushroom Toxicity & 30 &    The absence of odor and broad gill size suggest the mushroom is less likely to be poisonous, but the brown spore print indicates a higher risk of toxicity.          \\
PDF 1      & PDF Malware & 30 &    The PDF file is more likely to contain malware because it has a larger metadata size (262 KB), a larger total size (74 KB), and no Javascript keywords.             \\
PDF 2      & PDF Malware & 30 &  The large metadata size (180 KB), the large total size (7 KB), and fewer objects suggest the PDF contains malware.  \\
Student 1 & Student Performance & 30 & We believe this child is less likely to pass because they do not have family support, and we have seen that male students tend to be more likely to fail. \\
Student 2 & Student Performance & 30 & The lack of family support, and the sex (male) suggest the student is less likely to pass the class. But, the lack of a romantic relationship indicates an higher probability of passing. \\ \bottomrule
\end{tabular}
\end{table*}

\subsection{\Grader{}: Assessing the Quality of Narratives}
To assess the quality of narratives, we adapted metrics proposed in \cite{zhouEvaluatingQualityMachine2021}, which include:
\begin{itemize}
    \item \textit{Accuracy}: how accurate the narrative is to the original ML explanation.
    \item \textit{Completeness}: how much of the information from the original ML explanation is present in the narrative.
    \item \textit{Fluency}: how well the narrative matches the desired linguistic style and structure.
    \item \textit{Conciseness}: how long the narrative is considering the number of features in the explanation.
\end{itemize}

While there are other metrics that can be included, for the presentation of the \Grader, we focus on these four. The challenge with these metrics is that they are time consuming to manually compute, and manual assessment could introduce human bias into the evaluation.

\begin{figure*}[t]
    \centering
    
    
    
    
    
    \includegraphics{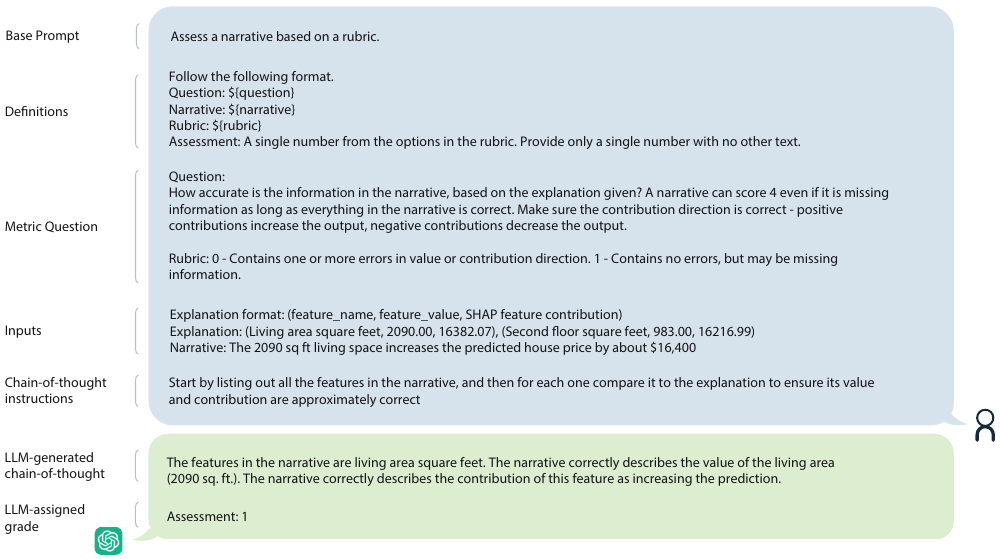}
    \caption{Prompt passed to the \Grader{} to compute the accuracy metric. Experiments suggested that asking for a grade from a 2-point rubric resulted in better performance than a simple yes/no question. 
    We saw that the \Grader{} incorrectly considered narratives that were accurate but did not include all feature values from the input explanation as inaccurate, and regularly did not notice when contribution directions were wrong if the values were correct. We added explicit instructions to the prompt for these two scenarios accordingly. Finally, we determined that using a chain-of-thought prompting approach further improved the \Grader{}'s ability to correctly score accuracy.
    }
    \label{fig:accuracy-prompt}
\end{figure*}

Our \Grader{} automatically evaluates the quality of the narrative by prompting an independent LLM, GPT-4o in our case, to consider the ML explanation and the generated narrative and assign a value to the respective metric. Fig. \ref{fig:accuracy-prompt} shows an example of a \Grader{} prompt evaluating the accuracy metric. In summary, it takes in the input and output of the \Narrator{} along with a rubric criteria, and comes up with a score depending on which metric it is asked to assess. In Fig.~\ref{fig:accuracy-prompt}, the \grader{} return will return either \textbf{1}, indicating an accurate narrative, or \textbf{0} for an inaccurate narrative.
Straight out-of-the-box, it can be difficult for LLMs to produce the exact desired output. 
Section \ref{sec:metrics} goes into details about how we tuned, validated, and used the \Grader{} system.

Putting effort into developing and validating a \Grader{} system offers several benefits. A well-validated grader can quickly adapt to a large variety of domains and dramatically reduce human-effort required to evaluate narratives. It can also be used as part of the finetuning process, as certain LLM optimization functions like bootstrapping few-shot examples optimize based on automated evaluation functions. Finally, and perhaps most importantly, a well-validated grader can be used as a guardrail in live deployment scenarios --- rather than risking showing inaccurate or low-quality narratives to users, a system can automatically revert to the default, graph-based ML explanation in situations where the generated narrative scores are unacceptable.

It is worth noting that in previous work, metrics such as METEOR~\cite{banerjee-lavie-2005-meteor} have been used to evaluate LLM output quality. However, to compute these metrics, we need a ground-truth label for every narrative. In real-world settings, we do not have these labels available, making it impractical in deployment. With \Grader, we alleviate this barrier.

\subsection{Using \Explingo{} for a New Application}
Having defined our two subsystems, we now describe the steps a user will take when applying \Explingo{} to a new ML decision-making problem\footnote{We have integrated this process in the Pyreal library, described in more detail in Section \ref{sec:pyreal}, which removes the manual effort required to generate and format explanations.}. We will consider a sample application of a user wishing to apply \Explingo{} to the task of house valuation. We assume the user has a ML model trained to predict house prices.

\begin{enumerate}
    \item The user generates a set of five SHAP explanations, explaining the model's prediction five (ideally diverse) different houses. For each of these explanations, the user assembles an \goldstandard{} by parsing them into the format described in the \textbf{Explanations} box of Fig. \ref{fig:narrator-prompt-simple}.
    \item For each \goldstandard{}, the user manually writes a short narrative describing the explanation in the desired style. For example, for an explanation of \textit{(neighborhood, Ocean View, 12039), (construction date, 2018, 3204)}, the user may write a narrative like \textit{This house is expected to sell for more than average due its prime location in Ocean View and because it was built fairly recently, in 2018.}
    \item The user adjusts the weighting of the \Grader{} metrics according to their needs; for example, they may choose to value accuracy higher than fluency. 
    \item The user can now pass explanations into the \Narrator{}, and get back narrative versions of these explanations. They can show these narratives to prospective homebuyers, real estate agents, or other stakeholders interested in understanding house prices.
    \item Optionally, the user may set up the guardrails process. In this case, they will pass the generated narratives into the \Grader{}. Any narrative that does not achieve some threshold score will be rejected and not shown to stakeholders.
\end{enumerate}

Other than potentially adjusting metric weighting, the user does not need to customize the \Grader{} for their application. Similarly, the only customization required for the \Narrator{} is to pass in a small set of \goldstandard{}s; the \Narrator{} automatically uses these narratives to generate narratives in the desired style that achieve high scores on the weighted metrics. We designed both subsystems to be application-agnostic and validated them on multiple diverse applications

We now dive deeper into the \Explingo{} framework.

\begin{table}[t]
\centering
\caption{Summary of metric validation datasets for the accuracy and completeness metrics. For accuracy, we include both complete and incomplete narratives to ensure they are both be rated as accurate.}
\label{tab:validation-datasets}
\begin{tabular}{@{}lp{4cm}cc@{}}
\toprule
Metric       & Type                                                    & Rubric Grade & Count \\ \midrule
Accuracy     & Accurate with all values                                       & 1            & 16                                 \\
             & Accurate with some values                                     & 1            & 9                                 \\
             & Accurate with approximate values                            & 1            & 9                                  \\
             & Error in values                                             & 0            & 8                                 \\
             & Error in contribution direction                             & 0            & 8                                 \\
            & Total & & 50 \\ \midrule
Completeness & \raggedright All features, values, and contributions                     & 2            & 10                                \\
             & \raggedright All features, values, and contribution directions           & 2            & 5                                   \\
             & \raggedright All features, but not all values or contribution directions & 1            & 17                                \\
             & Missing one or more features                                & 0            & 18                                 \\ 
             & Total & & 50 \\ \bottomrule
\end{tabular}
\end{table}

\section{\GoldStandard{} and Evaluation Datasets} \label{sec:datasets}
Narratives can be written in numerous ways and the desired style of narratives is highly subjective. Therefore, we introduce the concept of an \goldstandard{} dataset, which is a small dataset provided by the user containing high-quality examples of narratives. We steer both the \Grader{} fluency metric and the \Narrator{} based on these \goldstandard{}s.

For this work, we created nine datasets to take \goldstandard{}s from and to use for evaluations. We started by picking four publicly available ML datasets. We selected these datasets to include a wide variety of contexts ranging from predicting prices of houses to predicting toxicity of mushrooms. From these we created \goldstandard{}s. To create an exemplar dataset, we pick a data point and form an entry.  Each entry represents one model prediction on one data point. Each entry includes a SHAP explanation, an explanation format descriptor, and a brief context sentence describing what the model task is (e.g. predicting house prices). Our nine exemplar datasets are summarized in Table \ref{tab:datasets}. 

\begin{figure*}
    \centering







\includegraphics{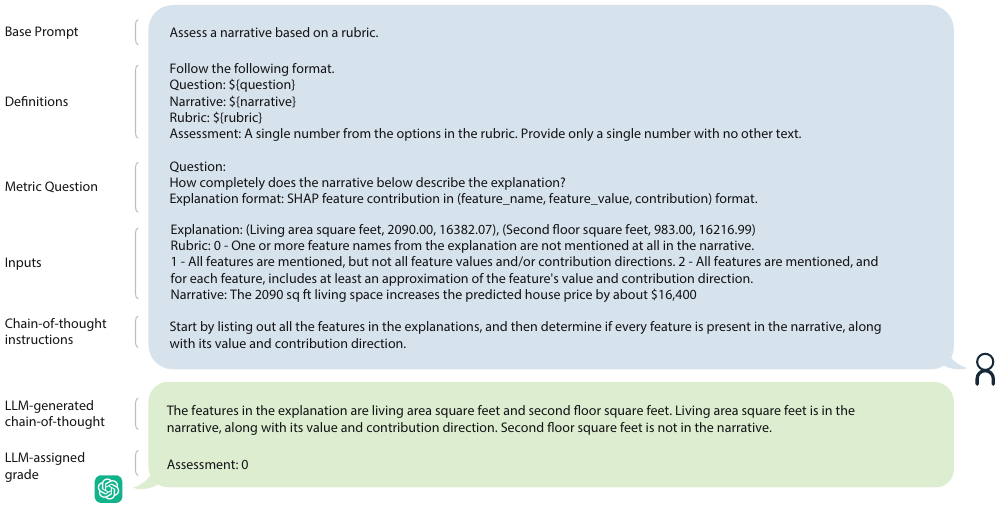}
    \caption{Prompt passed to the \Grader{} to compute the completeness metric. We found that while the original version of the prompt allowed the LLM to correctly identify missing feature values, it was regularly not identifying when features were missing altogether or did not have their feature directions listed. Through experimentation, we determined that a chain-of-thought \cite{wei_chain--thought_2023} approach that explicitly asked the \Grader{} to list out and consider each feature in the explanation one-by-one addressed this issue well.
    }
    \label{fig:completeness-prompt}
\end{figure*}

\textbf{Hand-written narratives:} For each \goldstandard{} dataset, we picked a few entries and created hand-written narratives given their SHAP explanations. These narratives were written by different authors and had a distinctive writing style. Different contexts (i.e., what model predicts) require some amount of customization to achieve the best possible narratives automatically from an LLM. These hand-written narratives thus become a foundational piece in our system. To use our system for a new context (a new model), the user creates a few entries and writes narratives with their preferred style and structure. These entries can then be used to optimize our system to provide better narratives specific for the context and users' preferred narration style.

\section{Automated Grading of Narratives} \label{sec:metrics}
In this paper, we focus on generating narratives that are optimized on four metrics: accuracy, completeness, fluency, and conciseness. 

Scoring conciseness is a straightforward computation based on word-count; the other metrics were graded using a GPT-4o LLM. We had the LLM grade each narrative 5 times and then return the average grade. To ensure consistency in grades, we would check and alert if fewer than 4 out of these 5 scores agreed; however, this alert was never raised during our experimentation. 

Our total grade $G$ for a narrative is then computed by weighting and aggregating all metrics with the following equation:
\begin{equation}
    G = \alpha_a A + \alpha_f F + \alpha_c C + \alpha_s S
\end{equation}
where $A$ is the Accuracy grade, $F$ is the Fluency grade, $C$ is the Completeness grade, $S$ is the conciseness grade, and the $\alpha$ parameters are weighting constants.

As described in more depth below, accuracy is scored on 2-point 0/1 scale, completeness is scored on a 3 point 0/1/2 scale, and fluency and conciseness are scored between 0 and 4. For our experiments, we therefore equally weighted all metrics by setting $\alpha_a=4, \alpha_c=2, \alpha_f=\alpha_s=1$, and report these weighted metric values in our results.

In running and optimizing LLMs, we used DSPy\footnote{https://github.com/stanfordnlp/dspy} \cite{khattab2023dspycompilingdeclarativelanguage}, an open-source LLM programming framework. DSPy enabled us to experiment with different LLM steering and tuning approaches and run evaluations in a structured manner. 


In the rest of this section, we introduce the metrics we used to evaluate narratives, and our process for developing and validating our \Grader{}.

\subsection{Accuracy} 
The accuracy metric evaluates if the information in the narrative is correct based on the input explanation. For example, in an accurate narrative, all feature values, SHAP feature contribution values, and  contribution directions (positive or negative) are correct. 
Narratives that do not mention all features can still be considered accurate. 
This metric was scored on a Boolean scale of 0 (inaccurate) or 1 (accurate).

Accuracy is scored by the \Grader{} by passing in a prompt to the LLM. This prompt includes the input ML explanation, the explanation format, and the narrative to grade, and asks the LLM whether or not the narrative was accurate.

In order to tune and validate the prompt passed into our \Grader{} for accuracy, we developed a metric validation dataset of 50 explanation-narrative pairs, along with human-labeled accuracy grades. These examples contained randomly selected SHAP explanations, with narratives. These narratives were intentionally written to span five levels of accuracy, as summarized in Table \ref{tab:validation-datasets}. 

We then started with a basic prompt (\textit{How accurate is the information based on the explanation given?}) and passed the dataset entries to the \Grader{}. We then compared the LLM-generated grades to the human-labeled grade for each entry to get a performance score across each level-of-accuracy category. Finally, we used these fine-grained results to manually adjust our prompt until we got sufficient performance (at least 95\% agreement with human labels). Fig. \ref{fig:accuracy-prompt} shows our final version of the prompt, as well as the manual adjustments we made.

\newcommand{\shadecell}[1]{\cellcolor[gray]{\ifnum#1>5 0.65\else\ifnum#1>2 0.9\else 0.95\fi\fi} #1}

\begin{table}[b]
\centering
\caption{Agreement between \Grader{} and human-labeled validation set for Accuracy (left) and Completeness (right).} \label{tab:accuracy_and_completeness}
\begin{tabular}{cccc}
   & & \multicolumn{2}{c}{\textbf{\Grader{}}} \\ 
  & & 0 & 1 \\ \midrule
  \multirow{3}{*}{\rotatebox{90}{\textbf{Human}}} & 0 & \shadecell{15} & \shadecell{1} \\
  & 1 & \shadecell{1} & \shadecell{33} \\ \\ \bottomrule
\end{tabular}
\hspace{1cm} 
\begin{tabular}{ccccc}
   & & \multicolumn{3}{c}{\textbf{\Grader{}}} \\ 
  & & 0 & 1 & 2 \\ \midrule
  \multirow{3}{*}{\rotatebox{90}{\textbf{Human}}} 
  & 0 & \shadecell{18} & \shadecell{0} & \shadecell{0} \\
  & 1 & \shadecell{0} & \shadecell{17} & \shadecell{0} \\
  & 2 & \shadecell{0} & \shadecell{2} & \shadecell{13} \\ \bottomrule
\end{tabular}

\end{table}

The final version of our \Grader{} agreed with human labels on 96\% of entries. A more detailed analysis of these results is shown in Table \ref{tab:accuracy_and_completeness}.

\begin{figure*}
\centering



% ... 




\includegraphics{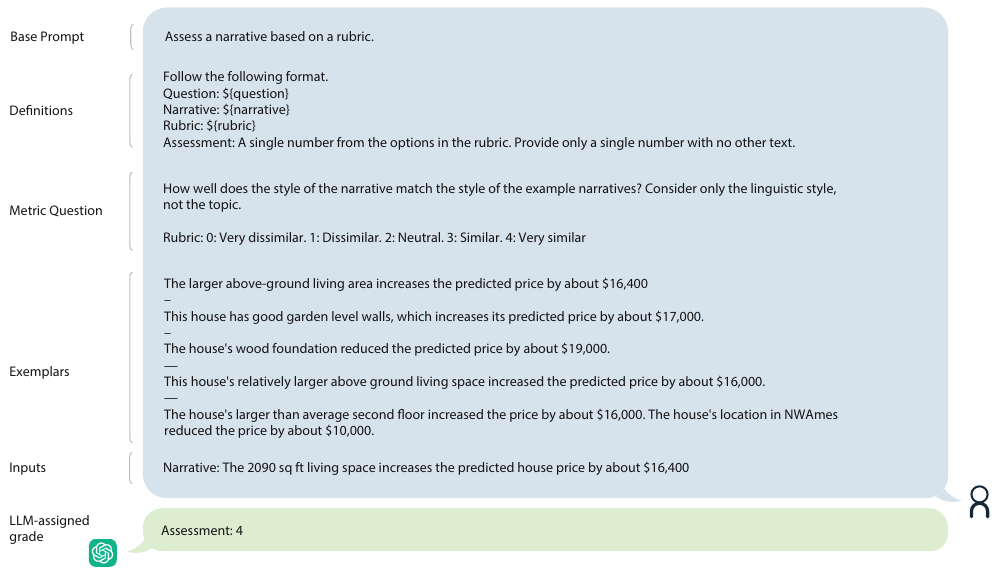}
    \caption{Prompt passed to the \Grader{} to compute the fluency metric. We found that the \Grader{} was over-emphasizing narrative content, as shown by a significant difference in score between narratives from different datasets with original prompt (\textit{How well does the narrative match the style of the example narratives?}). We therefore added a statement to explicitly ignore topic. Our experiments determined that there were diminishing returns to effectiveness after 5 \goldstandard{} narratives.}
    \label{fig:fluency-prompt}
\end{figure*}

\subsection{Completeness}
The completeness metric grades how much of the information from the input ML explanation is present in the narrative. For our experiments, we define a complete explanation as one that 1) mentions all features and their values from the ML explanation and 2) mentions the contribution direction (increasing or decreasing the model prediction) for all features. We note that this is subjective and context-dependent; for example, in some domains, the exact contribution values may be needed. Our prompts use a custom rubric that can be adjusted for different needs.

Completeness is scored by prompting an LLM with the explanation, explanation format, and narrative, and asking it to score completeness on a scale of 0 (missing one or more features), 1 (contains all features but missing one or more feature values or contribution directions) or 2 (includes all features, values, and contribution directions).

Like for the accuracy metric, we created a metric validation dataset of 50 human-labeled explanation-narrative pairs, as summarized in Table \ref{tab:validation-datasets}, that included narratives of different levels of completeness. 

Again, we started with a basic prompt (\textit{How completely does the narrative below describe the explanation given?}) and passed the dataset entries to the \Grader{}. We compared the LLM-generated grades to the human-labeled grade for each entry to get a performance score across each level-of-completeness category, and used these fine-grained results to manually adjust our prompt until we got sufficient performance (at least 95\% agreement with human labels). Fig. \ref{fig:completeness-prompt} shows our final version of the prompt, as well as the manual adjustments we made.

We ended up with the prompt shown in Fig. \ref{fig:completeness-prompt}, which we evaluated on our metric validation dataset. Using this prompt, the \Grader{}'s grades agreed with the human-labeled grades on 96\% of entries. The full results of validation are shown in Table \ref{tab:accuracy_and_completeness}.

\subsection{Fluency}
Fluency is the degree to which the narrative sounds natural or human-written. This is highly subjective and context-dependent; therefore, for this paper we grade fluency as how closely the narrative's linguistic style matches that of a few \goldstandard{} narratives. For example, consider the following two narrative styles for the same explanation:

\textbf{Precise:} The house's size (1203 sq. ft.) increased the ML output by \$10,203. The size of the second floor (0 sq. ft.) decreased the ML output by \$4,782.

\textbf{Casual:} The house's large size increases the price prediction by about \$10,200, while its lack of a second floor decreases the prediction by about \$4,700.

Users pass in a small set of hand-written \goldstandard{} narratives that are then used by the \Grader{} to grade fluency. We prompted the LLM with these \goldstandard{} narratives, along with the narrative to be graded, asking it to score how well the narrative matches the style of the \goldstandard{}s on a discrete scale from 0 to 4. Fig.~\ref{fig:fluency-prompt} shows the prompt we used to grade fluency.

To validate our fluency scoring function and identify the right number of exemplars to include in the prompt, we used our nine \goldstandard{} datasets introduced in Section \ref{sec:datasets}. From each dataset, we randomly selected $N$ hand-written narratives as \goldstandard{}s and asked the \Grader{} LLM to compare them to the remaining hand-written narratives. We then computed the mean and minimum difference between fluency scores on narratives evaluated against \goldstandard{}s from their own dataset and fluency scores on narratives evaluated against other datasets. 

\begin{figure}
    \centering
    \includegraphics[width=\linewidth]{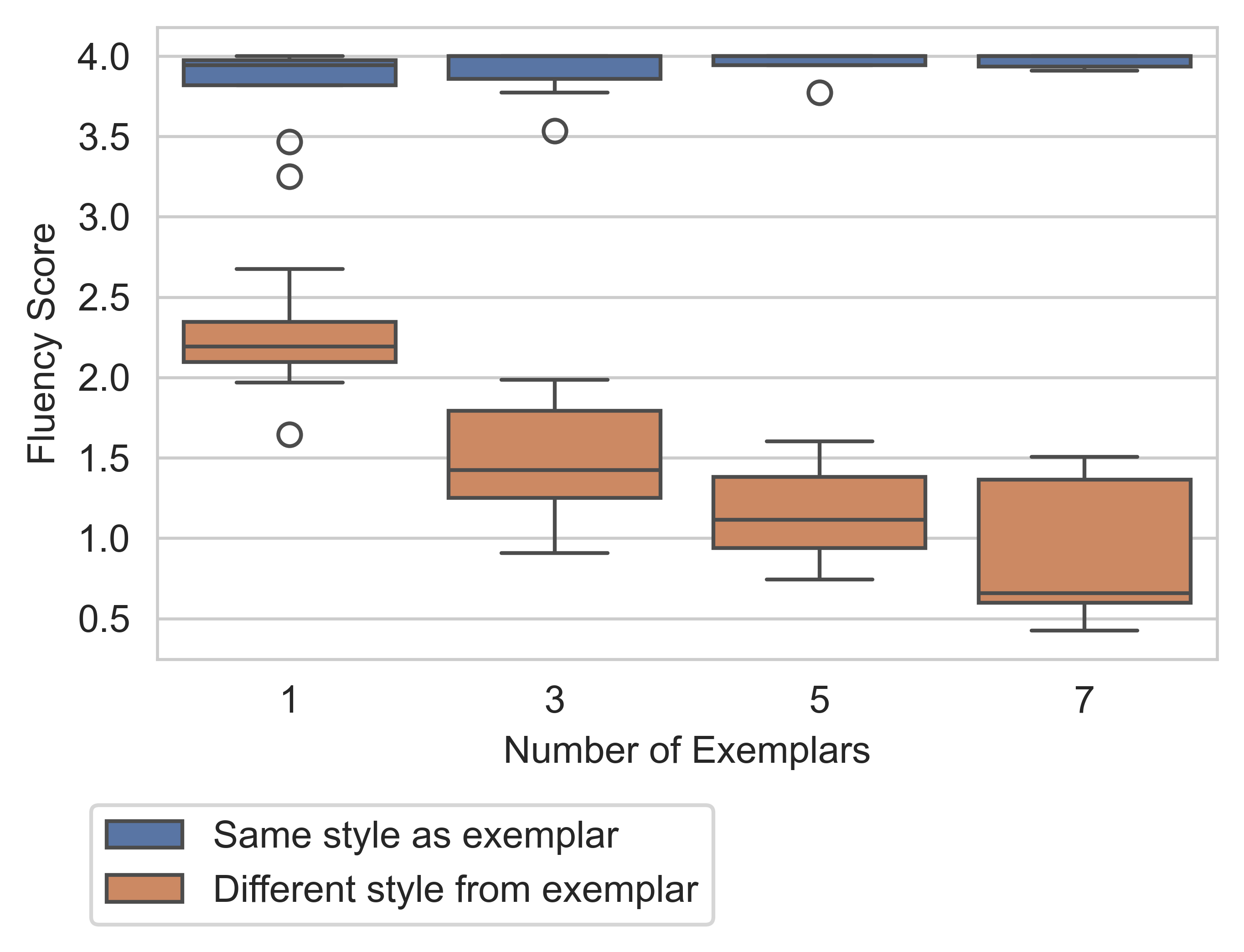}
    \caption{Difference between fluency scores on narratives compared to \goldstandard{}s from their own datasets (blue) and those compared against other datasets (orange). We expect to see significantly higher fluency scores for the former compared to the latter. We see that adding more \goldstandard{}s increases this difference, with diminishing returns after 5 \goldstandard{}s.}
    \label{fig:fluency_results}
\end{figure}

We compared results from $N=[1, 3, 5, 7]$. The results are visualized in Fig. \ref{fig:fluency_results}. We see that higher numbers of \goldstandard{}s increases the ability of the \Grader{} to differentiate narratives of the same style as the \goldstandard{}s from those in other styles, with diminishing returns after 5 \goldstandard{}s. We therefore set our \Grader{} to use 5 \goldstandard{}s (or as many as are available, if fewer than 5 are provided).

\subsection{Conciseness}
This metric grades how long the narrative is, scaled such that shorter narratives get higher grades. We believe that, all else being equal, users will tend to prefer shorter narratives. 

This metric is computed deterministically and scaled to a continuous value between 0 and 4, using the following equation:

\begin{equation}
\text{grade} = 
\begin{cases} 
  0 & \text{if } L \geq 2FL_{max} \\
  4 \times \left(2 - \frac{L}{FL_{\text{max}}}\right) & \text{if } FL_{max} < L < 2FL_{max} \\
  4 & \text{if } L \leq FL_{max}
\end{cases} 
\end{equation}
 where $L$ is the number of words in the narrative, $F$ is the number of features in the input explanation, and $L_{max}$ is some configured value representing the longest ideal number of words in the narrative per feature.

In effect, this equation grades a narrative as 4 if it is shorter than some hyperparameter-defined number of words per feature, 0 if it is longer than twice this value, with grades scaling linearly between these two extremes.

\section{Optimizing the Narrator} \label{sec:methods}
Having validated our metric scoring functions, we sought to improve our \Narrator{} to transform explanations into narratives that optimize these metrics. Like with our \Grader{} LLM, we used DSPy programming to assemble our prompts and used the GPT-4o API for our \Narrator{} LLM\footnote{We experimented with multiple larger API-based LLM models for the \Narrator{} including GPT-3.5, GPT-4o, Claude, Gemma, and Mistral models. Based on initial results, we decided to run experiments using GPT-4o.}.

As shown in Fig.~\ref{fig:narrator-prompt-simple} we have a base prompt that instructs the LLM what it needs to do given the three inputs: the explanation, the explanation format, and the context. Next, we show a few base prompts we tried. 

\textbf{Base Prompts}
We used the following prompts as a first step for baselines:

\begin{quote}
\textbf{Base Prompt 1:} You are helping users understand an ML model's prediction. Given an explanation and information about the model, convert the explanation into a human-readable narrative.
\newline

\textbf{Base Prompt 2:} You are helping users who do not have experience working with ML understand an ML model's prediction. Given an explanation and information about the model, convert the explanation into a human-readable narrative. Make your answers sound as natural as possible.
\newline

\textbf{Base Prompt 3:} You are helping users understand an ML model's prediction. Given an explanation and information about the model, convert the explanation into a human-readable narrative. Be sure to explicitly mention all values from the explanation in your response.
\end{quote}

We found that the \Narrator{} was able to generate reasonably complete and accurate narratives with these base prompts along with the the other three inputs. For the remaining experiments we only used base prompt 1, as it has the fewest tokens, scored highly on accuracy and completeness in our initial experiments, and does not include any information to suggest a specific style.

However, without exemplars of what a good narrative looks like for a specific application, the \Narrator{} is not able to customize the narrative style and length. For example, for an explanation from our House 1 dataset, the generated narrative using just the base prompt looks like \textit{``The model predicts the house price based on several key features. The above ground living area, which is 2090 square feet, contributes significantly to increasing the price by 16382.07 units.''} For this dataset, based on the hand-written \goldstandard{} narratives, we were seeking a narrative style more like \textit{``The relatively larger above-ground living space in this house increased its predicted price by about \$16,000.''}

As described in Section \ref{sec:datasets}, we selected five hand-written narratives for each \goldstandard{} dataset. We next consider if we can add these to the prompt to better customize narrative style.

\textbf{Hand-Written Few-Shot.} In this experiment, we add a small number (denoted as $H$) of \goldstandard{} hand-written narratives to the prompt. These narratives were randomly selected for each prompt from the hand-written narratives we have for the \goldstandard{} dataset. 

\textbf{Bootstrapped Few-Shot.} In the previous experiment, we are limited to the hand-written narratives available to us. Additionally, these exemplars may not be complete or accurate. We therefore used DSPy's \texttt{BootstrapFewShot} to create more exemplar narratives that score highly according to our \Grader{}. To do this we give the bootstrapper:
\begin{itemize}
    \item our hand-written narratives as a starting point;
    \item a requirement that exemplars it produces achieve a perfect score on accuracy, completeness, and fluency and at least 3.5 on conciseness as measured by our \Grader{}\footnote{We lowered the requirement slightly for conciseness as this significantly reduced bootstrapping time.}.
\end{itemize}

Once we have the bootstrapped exemplars, we add them, along with the the hand-written exemplars, to the prompt shown in Fig. \ref{fig:narrator-prompt-simple}. Thus the prompt now has $H$ hand-written and $B$ bootstrapped exemplars. With this addition, our prompt now has the base prompt, the explanation, the explanation format, the context, and $H+B$ exemplar narratives. 

\begin{table*}
\rowcolors{1}{white}{gray!10}
\centering
\caption{Narrative quality scores for narratives generated by our \Narrator{}, as evaluated by our \Grader{}, averaged across all datasets. Metrics are reported in terms of their weighted (0-4) values. Values greater than 3.5 are in \textcolor{blue}{blue}, values less than 3 are in \textcolor{red}{red}, and the highest scoring row for each metric is in \textbf{bold}. $H$=number of hand-written few-shot exemplars, $B$=number of bootstrapped few-shot exemplars.} \label{tab:results-by-technique}
\begin{tabular}{lcccccccc}
\toprule
Base Prompt & $H$ & $B$ & Accuracy & Completeness & Fluency & Conciseness & Total grade \\
\midrule
\midrule
Base Prompt 1 & 0 & 0 & \textbf{\textcolor{blue}{4.000 $\pm$ 0.00}} & \textbf{\textcolor{blue}{4.000 $\pm$ 0.00}} & \textcolor{red}{2.467 $\pm$ 0.76} & \textcolor{red}{0.850 $\pm$ 0.94} & \textcolor{blue}{11.317 $\pm$ 1.23} \\
Base Prompt 2 & 0 & 0 & \textcolor{blue}{3.911 $\pm$ 0.27} & \textbf{\textcolor{blue}{4.000 $\pm$ 0.00}} & \textcolor{red}{2.467 $\pm$ 0.66} & \textcolor{red}{0.872 $\pm$ 0.89} & \textcolor{blue}{11.250 $\pm$ 0.97} \\
Base Prompt 3 & 0 & 0 & \textbf{\textcolor{blue}{4.000 $\pm$ 0.00}} & \textbf{\textcolor{blue}{4.000 $\pm$ 0.00}} & \textcolor{red}{2.222 $\pm$ 0.82} & \textcolor{red}{1.056 $\pm$ 1.07} & \textcolor{blue}{11.278 $\pm$ 1.44} \\
Base Prompt 1 & 1 & 0 & 3.289 $\pm$ 1.35 & \textcolor{blue}{3.511 $\pm$ 0.56} & \textcolor{blue}{3.622 $\pm$ 0.47} & \textcolor{blue}{3.749 $\pm$ 0.14} & \textcolor{blue}{14.171 $\pm$ 1.67} \\
Base Prompt 1 & 3 & 0 & 3.111 $\pm$ 1.57 & 3.422 $\pm$ 0.53 & \textcolor{blue}{3.689 $\pm$ 0.41} & \textcolor{blue}{3.708 $\pm$ 0.19} & \textcolor{blue}{13.930 $\pm$ 1.82} \\
Base Prompt 1 & 5 & 0 & 3.111 $\pm$ 1.57 & 3.289 $\pm$ 0.56 & \textcolor{blue}{3.644 $\pm$ 0.49} & \textcolor{blue}{3.669 $\pm$ 0.35} & \textcolor{blue}{13.713 $\pm$ 2.08} \\
Base Prompt 1 & 1 & 1 & \textcolor{blue}{3.800 $\pm$ 0.40} & \textcolor{blue}{3.800 $\pm$ 0.23} & \textcolor{blue}{3.650 $\pm$ 0.34} & \textcolor{blue}{3.702 $\pm$ 0.32} & \textcolor{blue}{14.952 $\pm$ 0.58} \\
Base Prompt 1 & 1 & 3 & \textcolor{blue}{3.800 $\pm$ 0.40} & \textcolor{blue}{3.800 $\pm$ 0.23} & \textcolor{blue}{3.700 $\pm$ 0.26} & \textcolor{blue}{3.734 $\pm$ 0.28} & \textbf{\textcolor{blue}{15.034 $\pm$ 0.51}} \\
Base Prompt 1 & 3 & 1 & 3.323 $\pm$ 1.34 & \textcolor{blue}{3.538 $\pm$ 0.49} & \textcolor{blue}{3.846 $\pm$ 0.19} & \textcolor{blue}{3.843 $\pm$ 0.16} & \textcolor{blue}{14.551 $\pm$ 1.72} \\
Base Prompt 1 & 3 & 3 & 3.262 $\pm$ 1.33 & \textcolor{blue}{3.631 $\pm$ 0.58} & \textcolor{blue}{3.846 $\pm$ 0.23} & \textcolor{blue}{3.890 $\pm$ 0.11} & \textcolor{blue}{14.629 $\pm$ 1.73} \\
Base Prompt 1 & 5 & 1 & 3.289 $\pm$ 1.35 & 3.333 $\pm$ 0.63 & \textbf{\textcolor{blue}{3.933 $\pm$ 0.14}} & \textcolor{blue}{3.983 $\pm$ 0.03} & \textcolor{blue}{14.538 $\pm$ 1.63} \\
Base Prompt 1 & 5 & 3 & 3.289 $\pm$ 1.35 & 3.467 $\pm$ 0.66 & \textcolor{blue}{3.889 $\pm$ 0.27} & \textbf{\textcolor{blue}{3.988 $\pm$ 0.02}} & \textcolor{blue}{14.632 $\pm$ 1.65} \\
Base Prompt 1 & 5 & 5 & 3.378 $\pm$ 1.37 & 3.422 $\pm$ 0.64 & \textcolor{blue}{3.911 $\pm$ 0.15} & \textcolor{blue}{3.977 $\pm$ 0.03} & \textcolor{blue}{14.688 $\pm$ 1.66} \\
\bottomrule

\bottomrule
\end{tabular}
\end{table*}

\section{Results}

We now evaluate the quality of narratives generated by our \Narrator{}, and investigate how the exemplars provided influence these results.

\textbf{Experimental Setup.} Using the \Narrator{} we created narratives for explanations in the \goldstandard{} datasets. We used a variety of settings for the \narrator{}: 3 base prompts and 11 few-shot (exemplar) settings. In total, we had 169 explanations (across 9 datasets) and applied 13 techniques, resulting in 2,197 total narratives\footnote{The total LLM costs when using GPT-4o for generating and evaluating all narratives was about US\$8.44, or .38 cents per narrative to generate and evaluate on all metrics. The estimated number of input tokens required to transform one explanation ranges from about 150 to 1000 depending on explanation length and prompting technique.}. We scored each narrative on each of the four metrics described in Section \ref{sec:metrics}. For the conciseness metric, we set the maximum input length to 90\% of the longest feature description from the \goldstandard{} narratives. We also used these narratives (5 per dataset) to grade fluency. The mean metric grades from our evaluation by base-prompt/few-shot setting are shown in Table \ref{tab:results-by-technique}. 

\begin{figure}
    \centering
    \includegraphics[width=\linewidth]{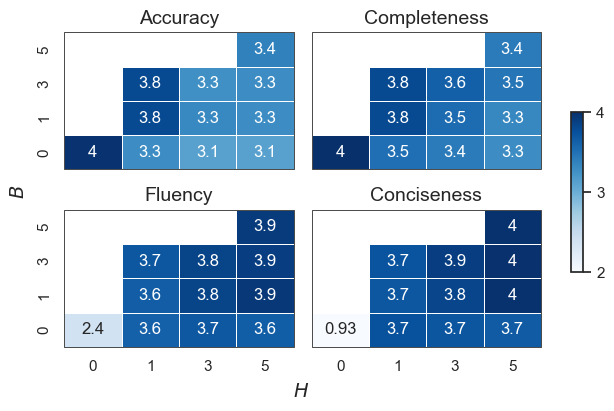}
    \caption{Correlation between mean metric grades (color) over all datasets by number of hand-written few-shot exemplars ($H$, x-axis) and number of bootstrapped few-shot exemplars ($B$, y-axis).}
    \label{fig:examples-vs-metrics}
\end{figure}

\textbf{Adding few-shot examples (exemplars) greatly improves narrative style at a small cost to correctness.}
In Fig. \ref{fig:examples-vs-metrics}, we visualize the impact that the number of exemplars has on each metric. We see a clear trend of fluency and conciseness grades increasing with more exemplars, while accuracy and completeness grades decrease. These findings are not surprising --- without an \goldstandard{}, the \Narrator{} has nothing to match to improve fluency and conciseness. On the other hand, more exemplars introduces more complexity that may increase the chance of the \Narrator{} hallucinating, and increase the chance of confusing the \Narrator{} with a potentially inaccurate \goldstandard{} narrative. The fact that we see high accuracy and completeness scores without any exemplars demonstrates that the LLM is able to complete the task with careful prompting out-of-the-box; adding exemplars mainly serves to adjust the style and structure of the narrative.
 
\textbf{We see the overall highest quality narratives with a small number of hand-written and bootstrapped exemplars.}
Our highest total grade across all four metrics was achieved with one hand-written few-shot exemplar and three bootstrapped exemplars. Having one of each style of few-shot exemplar achieved a close second-highest score.

\textbf{Bootstrapped exemplars contribute more to generating accurate and complete narratives.}
We also see a subtle trend in Table \ref{tab:results-by-technique} and Fig. \ref{fig:examples-vs-metrics} of bootstrapped exemplars improving accuracy and completeness over purely hand-written exemplars. This may be because bootstrapped exemplars must achieve perfect accuracy and completeness scores to be passed to the \Narrator{}, therefore removing the chance of passing in \goldstandard{} narratives that are either inaccurate or considered inaccurate by the \Grader{}.

\begin{table*}
\rowcolors{1}{white}{gray!10}
\centering
\caption{Narrative quality scores generated by our \Narrator{} in our two highest-scoring conditions ($H=1, B=1,3$), averaged by dataset. Values greater than 3.5 are in \textcolor{blue}{blue}, values less than 3 are in \textcolor{red}{red}.} \label{tab:results-by-dataset}
\begin{tabular}{lcccccc}
\toprule
Dataset & Accuracy & Completeness & Fluency & Conciseness & Total score \\
\midrule
House 1 & \textcolor{blue}{3.733 $\pm$ 0.40} & \textcolor{blue}{4.000 $\pm$ 0.00} & \textcolor{blue}{3.800 $\pm$ 0.30} & \textcolor{blue}{3.719 $\pm$ 0.24} & \textcolor{blue}{15.252 $\pm$ 0.58} \\
House 2 & \textcolor{blue}{4.000 $\pm$ 0.00} & \textcolor{blue}{4.000 $\pm$ 0.00} & \textcolor{blue}{3.911 $\pm$ 0.11} & \textcolor{blue}{3.836 $\pm$ 0.16} & \textcolor{blue}{15.748 $\pm$ 0.24} \\
House 3 & \textcolor{blue}{4.000 $\pm$ 0.00} & \textcolor{blue}{3.689 $\pm$ 0.27} & \textcolor{blue}{3.933 $\pm$ 0.10} & \textcolor{blue}{3.869 $\pm$ 0.14} & \textcolor{blue}{15.491 $\pm$ 0.35} \\
Mush 1 & \textcolor{blue}{3.556 $\pm$ 0.42} & \textcolor{blue}{3.600 $\pm$ 0.00} & \textcolor{blue}{3.511 $\pm$ 0.15} & \textcolor{blue}{4.000 $\pm$ 0.00} & \textcolor{blue}{14.667 $\pm$ 0.33} \\
Mush 2 & \textcolor{red}{1.760 $\pm$ 0.88} & \textcolor{red}{2.640 $\pm$ 0.36} & \textcolor{blue}{3.920 $\pm$ 0.18} & \textcolor{blue}{3.989 $\pm$ 0.03} & \textcolor{blue}{12.309 $\pm$ 0.67} \\
PDF 1 & \textcolor{blue}{4.000 $\pm$ 0.00} & \textcolor{red}{2.400 $\pm$ 0.00} & \textcolor{blue}{4.000 $\pm$ 0.00} & \textcolor{blue}{3.977 $\pm$ 0.03} & \textcolor{blue}{14.377 $\pm$ 0.03} \\
PDF 2 & \textcolor{red}{0.000 $\pm$ 0.00} & 3.040 $\pm$ 0.22 & \textcolor{blue}{3.840 $\pm$ 0.22} & \textcolor{blue}{3.949 $\pm$ 0.02} & \textcolor{blue}{10.829 $\pm$ 0.46} \\
Student 1 & \textcolor{blue}{4.000 $\pm$ 0.00} & \textcolor{blue}{3.600 $\pm$ 0.28} & \textcolor{blue}{3.960 $\pm$ 0.09} & \textcolor{blue}{4.000 $\pm$ 0.00} & \textcolor{blue}{15.560 $\pm$ 0.36} \\
Student 2 & \textcolor{blue}{3.840 $\pm$ 0.36} & \textcolor{blue}{3.920 $\pm$ 0.18} & \textcolor{blue}{4.000 $\pm$ 0.00} & \textcolor{blue}{3.880 $\pm$ 0.15} & \textcolor{blue}{15.640 $\pm$ 0.47} \\
\bottomrule
\end{tabular}
\end{table*}

For further investigation, we report the mean metric grades by dataset for the two highest-performing few-shot settings ($H=1, B=1/3$) in Table \ref{tab:results-by-dataset}. We see unexpectedly low scores for Mush 2 and PDF datasets. Inspecting the narratives generated more closely suggests a few possible issues.

\textbf{Some narrative styles require additional information to be generated and graded.}
The low accuracy for the PDF 2 dataset may stem from the \goldstandard{} narratives for this dataset using relative terms like ``the larger file size''; the \Grader{} does not get  information about the distribution of these features, so it may be labeling these terms as inaccurate. This suggests that if narratives with such comparative words are desired, additional information should be passed in.

\textbf{Further investigation into the \Grader{} quality on diverse datasets is required.} The low scores on the Mush 2 dataset and the lower completeness score on the PDF 1 dataset appear to stem from an error with the \Grader{} rather than the \Narrator{}. For example, for the Mush 2 dataset, the narrative \textit{The lack of odor and broad gill size suggest the mushroom is more likely to be poisonous} from the explanation \textit{(odor, none, 0.15), (gill-size, broad, 0.06)} was incorrectly given a 0 on accuracy. It is possible that the LLM is ``playing it safe'', given the high-stakes application of mushroom toxicity, as a result of AI safety work (this problem may not affect Mush 1, as its narratives include safety-conscious additions like ``be careful'' or ``confirm with external sources''). However, this is speculation that requires further investigation.

In the next section, we discuss our key takeaways from these results further.

\section{Discussion and Future Work}
This work represents a first step into transforming ML explanations into narratives, and suggests several future directions. We determined that powerful modern LLMs are capable of creating high-quality narrative versions of SHAP explanations.
Balancing hand-written and bootstrapped exemplars, while keeping the number of these low, appears to create the best balance between correctness (accuracy and completeness) and style (fluency and conciseness). Adding additional \goldstandard{}s can improve style at the cost of correctness --- users in different domains may value these metrics differently, and wish to modify the number of \goldstandard{}s accordingly. 

However, certain subtle issues --- such as including comparative terms like ``larger'' or working within complex domains --- confuse our \Grader{} into scoring narratives incorrectly. Without a highly-accurate grader, it may be difficult to generate narratives with enough trust for high-risk domains. Therefore, further investigation is needed.

\subsection{System Integration and Open-Source Implementation} \label{sec:pyreal}
We integrated the findings from this paper in Pyreal, a library for transforming explanations for readability. We implemented the \Narrator{} in the form of the \texttt{NarrativeTransformer}\footnote{Instructions for transforming explanations using Pyreal can be found here: https://dtail.gitbook.io/pyreal/user-guides/using-applications-realapps/narrative-explanations-using-llms} 
object in this library, an object that transforms explanations into narratives given some hyperparameters including style samples.

Pyreal can be used to both generate the initial explanation and then transform it into narratives. Fig. \ref{fig:pyreal-code} shows a sample of the code to generate a SHAP feature contribution explanation and transform it to narrative format.

\subsection{Lessons from Adjusting the \Grader{}}
In past work \cite{zytek_llms_2024}, we suggested that the accuracy metric be graded on a 3-point scale of \textit{inaccurate}, \textit{accurate but misleading}, or \textit{accurate}. However, discussion and experimentation revealed that ``misleading'' in this context is a highly subjective specifier, which makes it difficult for both human and LLM graders to give consistent grades. We therefore settled on a boolean 0 or 1 grade, which was sufficient for our purposes.

We originally intended for the fluency metric to broadly describe the degree to which the narrative sounds natural or human-written, fine-tuned from a general-purpose crowd-generated \goldstandard{} dataset of narratives on a wide variety of datasets. However, we determined from discussion with diverse ML users that preferred narrative style is highly subjective and context-dependent, and therefore the fluency metric would be far more useful if based on a small, application-specific \goldstandard{} dataset.

\begin{figure}[t!]
   \centering
   \begin{lstlisting}[language=Python]]
from pyreal.transformers import NarrativeTransformer
from pyreal import RealApp
    
context = "The model predicts house prices"
    
# Explanation, narrative pairs
examples = {"feature_contributions": 
        [("size, 120, -120", "The model...")]}
                
narrative = NarrativeTransformer(
               num_features=3,
               context_description=context,
               training_examples=examples)

app = RealApp(model, transformers=[narrative])   
app.produce_feature_contributions(X, X_train)\end{lstlisting}
    \caption{Open-source library implementation of narrative generation process. The output of the final line is a dictionary with keys as row IDs in X and values as narratives of that model prediction on that row.}
    \label{fig:pyreal-code}
\end{figure}

\subsection{Future Work}
In this work we focused on SHAP contributions; however, by adjusting the input explanation format and some details in the metric rubrics, this work could easily be updated to support other kinds of explanations.

While in this paper we focus on automated grading, a natural next step would be to evaluate how effective the narratives are for real-world decision making. 

It is essential to emphasize that this work represents only a first step toward enabling explanations that offer deeper insights and greater clarity. Our ultimate goal is to develop a system that not only generates understandable narratives but also enhances users' ability to make informed decisions based on those narratives. This goal requires future research focused on providing richer context and deeper insights in the explanations, allowing users to engage more critically with the information presented. One example of such insights includes rationalizing model behavior; for instance, an explanation narrative of ``This house is predicted to sell for more because it was built in the year 1930'' can be rationalized as ``In this market, older houses hold a premium because of their better building materials and unique history.'' In cases where the model uncovered unexpected patterns in data, such rationalizations can be valuable for knowledge discovery and decision-making.

Additionally, we encountered some challenges where our system could not generate and correctly grade comparative narratives (such as referring to a file as ``larger'' or ``smaller'') because we did not pass in the necessary context information.
To address these issues, we plan to explore methods for clearly defining data context. For instance, we could pass statistical benchmarks such as mean feature values or more detailed feature distributions to the LLMs.
Developing explanations that correctly use comparative terms may improve narrative fluency, and 
help users understand why certain characteristics, such as file size, are indicative of potential outcomes, ultimately enriching the narrative quality.

We determined that finetuning approaches are not required to develop an effective \Narrator{} with a sufficiently powerful base LLM. However, finetuning may be beneficial to allow smaller architectures to perform the task as well as the larger models we worked with. 
As a first step to trying this, we applied the same final prompt (with 3 labeled few-shot exemplars) to a smaller, 7B parameter local model (Mistral-7B-v0.1) to determine how well it could perform out-of-the-box. The model effectively matched style, with average fluency and conciseness scores of 3.91 and 3.81, respectively. However, the small \Narrator{} did not always parse the input explanation correctly, resulting in lower accuracy and completeness scores of 0.98 and 1.64 respectively. Since smaller, local models are more practical for many real-world applications, we plan to refine prompts and explore finetuning to improve these results.

\section{Conclusion}

This work presents findings in transforming traditional ML explanations into natural-language narratives using LLMs. We developed \Explingo{}, a system for generating narrative versions of explanations. This system includes a \Narrator{} that uses LLM's to transform ML explanations into narratives, and a \Grader{} that automatically grades narratives on metrics including accuracy, completeness, fluency, and conciseness.

Our experiments show that the \Narrator{}, when properly guided by a small \goldstandard{} dataset, can generate narratives that score highly on these metrics. We found that incorporating a mix of hand-written and bootstrapped few-shot exemplars improves narrative quality. 

This work represents a step towards enabling natural question-and-answer conversations between ML models and humans. Future research will continue this thread through  exploring additional types of explanations and conducting user studies to validate the effectiveness of narrative versions of explanations in real-world applications.





\bibliographystyle{IEEEtran}
\bibliography{IEEEabrv,references.bib}
%

\end{document}